\documentclass[letterpaper, 10 pt, conference]{ieeeconf}  

\IEEEoverridecommandlockouts                              

\usepackage{graphics} 
\usepackage{epsfig} 
\usepackage{times} 
\usepackage{amsmath} 
\usepackage{amssymb}  
\usepackage{booktabs}
\usepackage{caption}
\usepackage{graphicx}
\usepackage{float} 
\usepackage{subcaption}
\usepackage{color}
\usepackage{multirow}

\title{\LARGE \bf
$\nu$-DBA: Neural Implicit Dense Bundle Adjustment\\Enables Image-Only Driving Scene Reconstruction
}

\author{Yunxuan Mao$^{1}$, Bingqi Shen$^{1}$, Yifei Yang$^{1}$, Kai Wang$^{2}$, Rong Xiong$^{1}$, Yiyi Liao$^{1}$, and Yue Wang$^{1*}$
\thanks{$^{1}$Yunxuan Mao, Bingqi Shen, Yifei Yang, Yiyi Liao, Rong Xiong, and Yue Wang are with Zhejiang University, Hangzhou, China. }%
\thanks{$^{2}$Kai Wang is with the Application Innovate Lab, Huawei Incorporated Company, Beijing, China.}
\thanks{*Corresponding author.}
}

\begin{document}

\maketitle
\thispagestyle{empty}
\pagestyle{empty}


\begin{abstract}
The joint optimization of the sensor trajectory and 3D map is a crucial characteristic of bundle adjustment (BA), essential for autonomous driving. This paper presents $\nu$-DBA, a novel framework implementing geometric dense bundle adjustment (DBA) using 3D neural implicit surfaces for map parametrization, which optimizes both the map surface and trajectory poses
using geometric error guided by dense optical flow prediction. 
Additionally, we fine-tune the optical flow model with per-scene self-supervision to further improve the quality of the dense mapping. Our experimental results on multiple driving scene datasets demonstrate that our method achieves superior trajectory optimization and dense reconstruction accuracy. We also investigate the influences of photometric error and different neural geometric priors on the performance of surface reconstruction and novel view synthesis. Our method stands as a significant step towards leveraging neural implicit representations in dense bundle adjustment for more accurate trajectories and detailed environmental mapping.

\end{abstract}


\section{Introduction}

Building a dense map of the driving scene is desirable for autonomous vehicles, aiding localization, planning, and simulation. Bundle adjustment (BA) is indispensable for this reconstruction task when using images with noisy trajectories, which jointly optimizes camera poses and map points. However, the sparse images and small parallax angles along the driving route make dense surface mapping challenging using only images and noisy trajectories in driving scenes.



\begin{figure}[t]
\centering
\includegraphics[width=\linewidth,trim=4 4 4 4,clip]{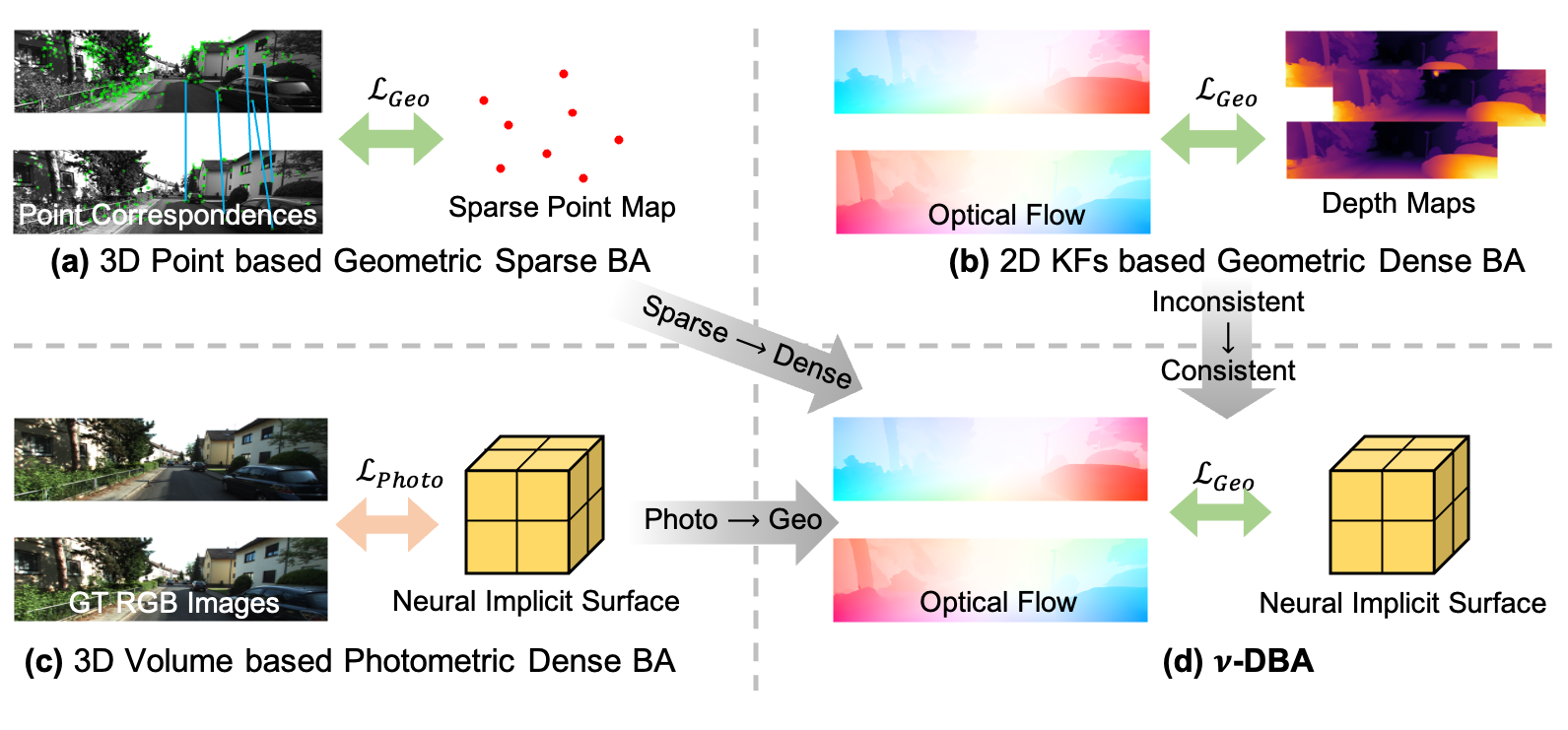}
\caption{\textbf{Comparison of BA methods.} (a) Traditional geometric sparse BA with 3D point map. (b) Geometric dense BA with 2D keyframes. (c) Photometric dense BA with the neural implicit surface. (d) \textbf{$\mathbf{\nu}$-DBA}: geometric dense BA with the neural implicit surface.
} 
\label{teaser}
\end{figure}

To investigate this problem, we delve into BA from the perspectives of \textit{map parametrization} and \textit{measurement metric}, respectively. 
The most popular BA methods take sparse 3D scene points cloud as the map parametrization, as shown in Fig. \ref{teaser} (a). This representation is consistent across different views, yet falls short in describing the dense geometry. Nevertheless, these sparse BA methods provide insights into the measurement metric. There are two types of measurement metrics in sparse BA:
geometric error and photometric error. COLMAP \cite{schoenberger2016sfm} and photometric BA \cite{alismail2017photometric} are two representative works, respectively. 
Geometric BA utilizes the scene point correspondence across images for optimization, while photometric BA utilizes the local patch consistency of the scene point projections. In \cite{campos2021orb}, geometric BA is shown to be more accurate than photometric BA, which might be induced by sensitivity in illumination and calibration.

In recent years, dense BA has received increasing attention. A representative work, DROID-SLAM~\cite{teed2021droid}, parametrizes the dense map based on a set of 2D keyframes and their optimizable depth maps, as shown in Fig. \ref{teaser} (b). A dense map can be obtained by projecting the depth maps to 3D. However, keyframe depth might be redundant in 3D, and there are no 3D consistency constraints to enforce the overlap part to be the same, which may limit the accuracy of the resultant map. Despite the limitation in the map parameterization, DROID-SLAM inherits the advantage of geometric BA, as it can be essentially considered as geometric BA by employing dense optical flow using RAFT~\cite{teed2020raft} as the measurement. This makes DROID-SLAM a geometric dense BA (DBA), as the scene points correspondence in geometric sparse BA is akin to sparse optical flow.

Recent advances in neural implicit representation have achieved impressive high-fidelity novel view synthesis~\cite{mildenhall2021nerf} and scene reconstruction~\cite{wang2021neus, Yu2022MonoSDF}. With the differentiable rendering, the image regression loss can be back-propagated to learn the 3D representation~\cite{wang2021neus, Yu2022MonoSDF}.  Considering such neural mapping from the perspective of dense BA, we find that the neural implicit representation acts as a 3D dense map parametrization, avoiding the consistency issue of keyframe depth maps. The differentiable rendering can be viewed as a photometric error-based measurement model,
yielding a photometric dense BA, as shown in Fig. \ref{teaser} (c). Utilizing such a photometric dense BA as the back-end, several neural implicit representation-based SLAM systems are proposed~\cite{zhu2022nice, zhu2023nicer, johari2022eslam, mao2023ngel, zhang2023goslam}. Despite some of them are applicable to RGB images for indoor scenes~\cite{zhu2023nicer,zhang2023goslam}, existing methods in this direction struggle in street scenes with RGB input, indicating that street scenes pose challenges due to complex illumination and textureless regions. This raises a question: \textit{Can we bridge the 3D neural implicit representation with the geometric error to improve the dense BA accuracy?}



Following the idea, in this paper, we propose $\nu$-DBA (pronounced as ``neuDBA''), a geometric dense BA framework with 3D neural implicit surface as map parametrization, which optimizes both the map surface and trajectory poses using geometric error derived from dense flow, as illustrated in Fig. \ref{teaser} (d). 
Apart from the framework, we investigate the effectiveness of photometric error and other neural geometric prior on the quality of the surface reconstruction and the novel view synthesis. Moreover, to relieve the generalization gap of the flow prior, we further propose to fine-tune the flow model through per-scene self-supervision. In the experiments, we evaluate our method on several driving scene datasets and demonstrate superior trajectory and mapping accuracy. In summary, we make the following contributions:
\begin{itemize}
\item We propose a geometric error-based dense BA framework with 3D consistent representation i.e. neural implicit surface as the map parametrization, for driving scene reconstruction.

\item We propose to self-supervise the optical flow model to narrow the generalization gap, which further improves the quality of the reconstructed surface.


\item We evaluate the framework on several driving scene datasets, demonstrating superior performance in reconstruction and trajectory optimization, and investigating the effect of photometric error and other neural geometric priors.
\end{itemize}



\section{Related Works}
\subsection{Bundle Adjustment}
A comprehensive and precise road map is key for the effective operation of autonomous vehicles, facilitating their capabilities in self-localization, trajectory planning, and simulation exercises.
Bundle Adjustment (BA) plays a critical role in this mapping process, especially when dealing with imprecise trajectories and imagery, by simultaneously adjusting the estimations of camera orientations and landmark locations. We conduct an in-depth exploration of Bundle Adjustment (BA) with respect to the dimensions of map parameterization and measurement metrics. The measurement metrics of existing mature BA include two main types: geometric error and photometric error. 
Methods that use geometric error perform optimization in two steps. First, the raw sensor measurements are pre-processed to generate an intermediate representation, solving part of the overall problem, such as computing the image coordinates of corresponding points. Second, the computed intermediate values are used to estimate the geometry and sensor poses. 
Photometric error-based methods skip the pre-computation step, directly using the actual sensor data for optimization.

Photometric error-based methods, such as DSO \cite{2016Direct}, LSD-SLAM \cite{engel2014lsd}, and DTAM \cite{newcombe2011dtam}, rely on minimizing the photometric error between consecutive frames to estimate the camera pose and reconstruct the scene structure. These methods suffer from challenges like illumination changes and geometric distortions, such as rolling shutter artifacts, which can adversely affect their performance.
In contrast, geometric error-based methods, such as MonoSLAM \cite{davison2007monoslam}, PTAM \cite{klein2007parallel}, and ORB-SLAM3 \cite{campos2021orb}, solve the problems of direct methods by generating features and estimating a geometry prior before bundle adjustment. However, in early works, both methods utilize sparse 3D feature point cloud as the map parametrization due to the sparse feature correspondence across frames. The sparse bundle adjustment is limited when lacking features. 

With the development of hardware, dense mapping has received increasing attention. DROID-SLAM \cite{teed2021droid} proposed a geometric dense BA with dense optical flow. DROID-SLAM builds a dense 3D map by projecting the estimated depth of 2D keyframes to 3D. might be redundant in 3D, and there are no 3D consistency constraints. In this paper, we aim to bridge the 3D neural implicit representation with the geometric error to improve the dense BA.

\subsection{Neural Implicit Scene Representation} 

Neural implicit representations have gained popularity in 3D reconstruction \cite{park2019deepsdf, mescheder2019occupancy, sitzmann2020implicit, oechsle2021unisurf, chen2021mvsnerf, wang2022go, yu2023nf}. As dense, consistent scene representations, these methods can extract high-fidelity surfaces useful for autonomous driving tasks. At the same time, neural radiance field (NeRF) \cite{mildenhall2021nerf} has achieved impressive novel view synthesis results with volume rendering techniques. The differentiable volume rendering techniques make it possible to optimize neural implicit surface representation with 2D information \cite{wang2021neus, long2022sparseneus}. From the dense BA perspective, we find that the neural implicit representation acts as a 3D map parametrization, and the differentiable rendering acts as the measurement model.

Recently, building upon such photometric dense BA, several neural implicit representation-based SLAM systems have been proposed \cite{zhu2022nice, sucar2021imap, wang2023coslam, zhu2023nicer, zhang2023goslam, mao2023ngel}. These works enable dense BA with neural implicit representation. However, most of them focus primarily on indoor scenes with dense viewpoints and utilize the ground truth depth image for optimization tasks. Yet, ground truth geometric priors, such as depth images or LiDAR, are not available for all driving scenes.
NICER-SLAM \cite{zhu2023nicer} proposes a ground truth geometric prior-free dense SLAM system for indoor scenes but does not extend to outdoor unbounded scenes with driving views. The viewpoints are forward-facing with long and narrow trajectories in the outdoor unbounded driving scenes, which differ significantly from indoor scenes. StreetSurf \cite{guo2023streetsurf} proposes a multi-shell cuboid neural scene representation for street driving views. However, the monocular geometry cues used in StreetSurf cannot be self-supervised and are therefore constrained by the training dataset. By contrast, we propose a geometric dense BA framework with a 3D neural implicit surface as the map parametrization for RGB input in outdoor driving scenes and fine-tune the optical flow model for each scene through self-supervision to enhance performance.


\begin{figure*}[t]
\centering
\includegraphics[width=\linewidth,trim=4 4 4 4,clip]{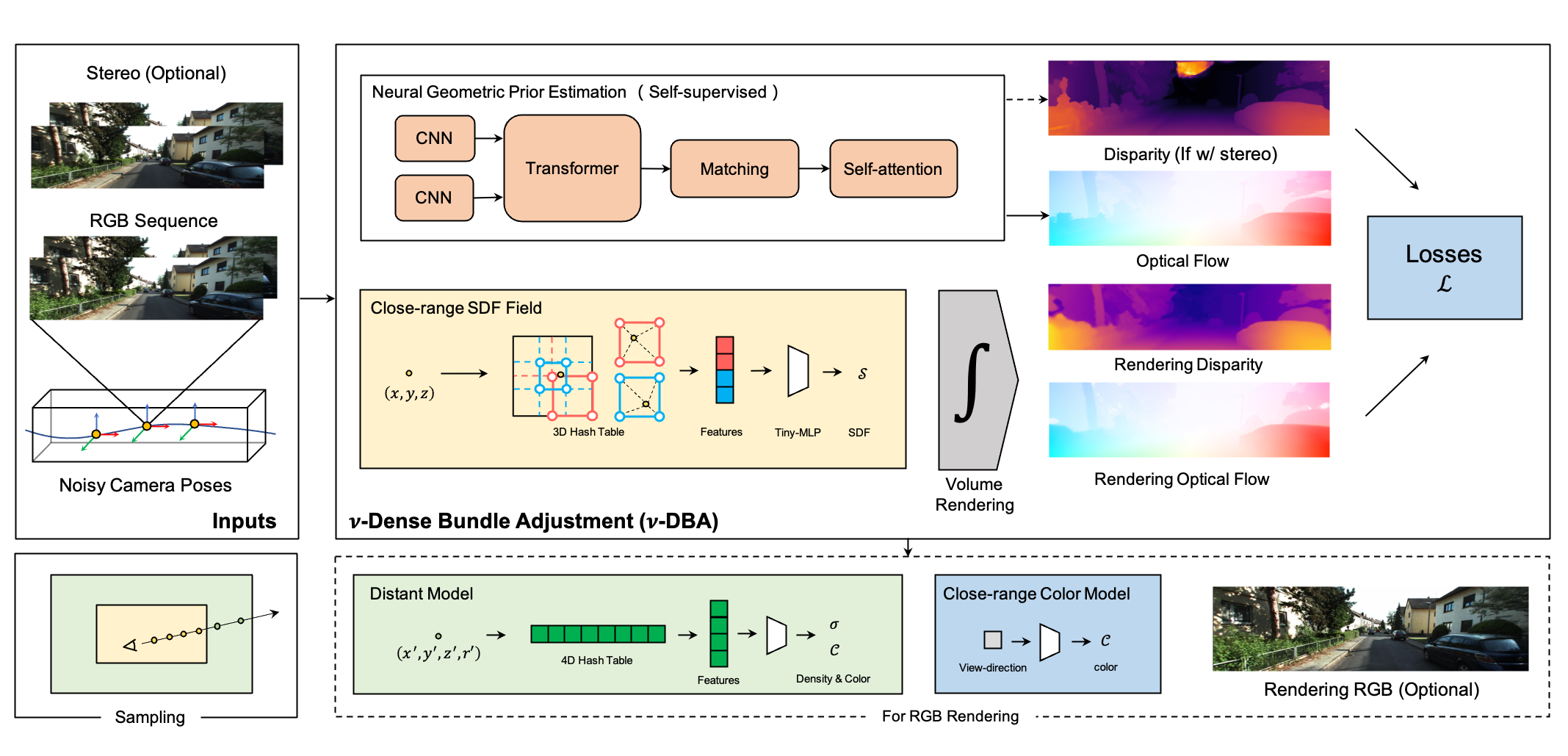}
\caption{\textbf{Overview.} Our method takes a sequence of RGB images with noisy camera poses as input. In $\nu$-DBA, the self-supervised model inference optical flow and disparity as neural geometric priors, which is utilized to supervise the close-range SDF field and the camera poses with geometric losses. Moreover, a distant model and a close-range color model are used for RGB rendering.
} 
\label{overview}
\end{figure*}

\section{$\nu$-Dense Bundle Adjustment}

Given a sequence of RGB images $\mathbf{I} = \{I_0, I_1, \dots, I_k\}$ with noisy camera poses $\mathbf{T}=\{T_1, T_2,\dots, T_k\}$, the proposed $\nu$-DBA is capable of simultaneously extracting dense and complete surfaces and optimizing the camera poses. As shown in Fig. \ref{overview}, there are two core components: the close-range SDF field as map parameterization, and the flow loss as the geometric metric. In addition, a self-supervision method is employed to narrow down the generalization gap of off-the-shelf optical flow predictor.


\subsection{Close-Range SDF Field}
The close-range model represents geometry as a signed distance function (SDF). Given a 3D point, it returns the point's distance to the closest surface. In this work, the SDF function is parameterized by a 3D hash table $h_\theta$ as the feature grid, and a single MLP $f_\theta$ as the geometry decoder
\begin{equation}
\hat{s} = f_\theta(h_\theta(\mathbf{x}))
\label{close sdf model}
\end{equation}
Here, $\mathbf{x}$ is the 3D point, $\hat{s}$ denotes the corresponding SDF value, and $\theta$ are learnable parameters.

\subsection{Neural Geometric Prior with Self-Supervision} \label{cge}

To build the geometric error, we choose optical flow estimated by a pre-trained model as the neural geometric prior i.e. observations. When a stereo pair is available, disparity is also employed.

\subsubsection{Optical Flow}
Optical flow finds the 2D-pixel displacement field between two images, which can be easily obtained using an off-the-shelf flow estimator. We use the pretrained unimatch model \cite{xu2023unifying} to estimate the optical flow $\mathbf{F}_i$ for each input frame pair $\{I_{i}, I_{i+1}\}$. Compared with monocular depth, the optical flow infers the correspondences, which is not affected by the metric scale.

\subsubsection{Stereo Matching}
Stereo matching can be regarded as correspondence between the left and right cameras, of which the relative pose is known, thus reducing the correspondence space to epipolar lines. The unimatch model also estimates disparity $D_{disp}$ for a given stereo pair $I^l_{i}, I^r_{i}$. The accuracy of stereo matching is often higher than that of optical flow, but it is applicable only in scenes where a stereo pair is available.

\subsubsection{Self-supervised Fine-tune}

One benefit of flow is that the model can be self-supervised using only RGB image inputs, which is able to narrow down the generalization gap of the optical flow inference model. Therefore, we follow \cite{long2022sparseneus} to fine-tune the inference result of the optical flow model in a per-scene manner. Similar to \cite{jonschkowski2020matters}, we apply census loss, smoothness loss, and self-supervision loss to the optical flow model. 

\subsubsection{Other Neural Geometric Priors}

Following \cite{Yu2022MonoSDF, guo2023streetsurf}, the proposed model can also be supervised by the monocular geometry cues i.e. monocular depth and normal. Note that compared with flow, the monocular depth and normal can only be self-supervised when camera poses are given, 
which may limit the generalization performance of monocular cues. In addition, we consider that noise in flow prediction is more uniform in near and far regions than that in monocular depth and normal prediction, which makes flow more aligned with the uniform weighting in loss terms.

\subsection{Optimization}
The $\nu$-DBA optimizes the SDF field and the camera poses together with the supervision of 2D neural geometric prior estimated by the model in Section~\ref{cge}.

\subsubsection{Volume Rendering} \label{close vr}
Following recent work \cite{mildenhall2021nerf}, we employ volume rendering as a measurement model, to learn the pose and map supervised by estimated optical flow. Given the camera's intrinsic parameters and current camera pose, we cast a ray $\mathbf{r}$ from the camera center $\mathbf{o}$ through the pixel along its normalized view direction $\mathbf{v}$. We sample $N$ points along a ray, denoted as $\mathbf{x}_i = \mathbf{o} + d_i\mathbf{v}$, where $d_i$ is the depth of point $\mathbf{x}_i$, and $i \in {1,\cdots, N}$.

Given an image pair $\{I_k,I_j\}$ and a ray $\mathbf{r}_j$ from the pixel $\mathbf{p}_j$ in image $I_j$, we can calculate the pixel location of every sample $\mathbf{x}^{\mathbf{r}_j}_i$ in the ray $\mathbf{r}_j$ on another image via differentiable tomography as in \cite{yao2018mvsnet}

\begin{equation}
H(d)=K_kR_k\left(I-\frac{(R_k^{-1} \mathbf{t}_k-R_j^{-1}\mathbf{t}_j)\mathbf{n}^\top_jR_j}{d}\right)R_j^{-1}K_j^{-1}
\label{homo}
\end{equation}
that $\{K_j, R_j, \mathbf{t}_j\}$ are the camera intrinsics, rotations, and translations of frames $\{I_j\}$, $\mathbf{n}$ is the principle axis of the reference camera, and $d$ is the depth of the point in reference view $I_j$. Then, we can get the pixel location of sample $\mathbf{x}^{\mathbf{r}_j}_i$ on image $I_k$: $\mathbf{p}^i_k \sim H(d_i)\cdot\mathbf{p}_j$ where '$\sim$' denotes the projective equality. 

For volume rendering, following NeuS \cite{wang2021neus}, we first query the SDF value $\hat{s}_i$ of sample $\mathbf{x}_i$ and convert the SDF values into density values ${\sigma_i}$. The optical flow $\hat{\mathbf{F}}$ of pixel $\mathbf{p}_j$ from frame $I_j$ to frame $I_k$ can be calculated as 
\begin{equation}
\hat{\mathbf{F}}(\mathbf{r}_j) = \sum_{i=1}^N T_i\alpha_i \mathbf{p}^i_k - \mathbf{p}_j
\label{render flow}
\end{equation}
where
\begin{equation}
  T_i = \prod_{j=1}^i(1-\alpha_j) ~ \text{and} ~ \alpha_i = 1-\exp(-\sigma_i\delta_i)
\label{render}
\end{equation}

Similarly to optical flow, we can render the disparity when given a stereo pair. We can calculate the disparity of a sample ray $\mathbf{r}$

\begin{equation}
\hat{D}_{disp}(\mathbf{r}) = \sum_{i=1}^N T_i\alpha_i \frac{f\cdot b}{d_i}
\label{disp}
\end{equation}
where $f$ is the focal length of the camera and $b$ is the baseline of the stereo pair.

\subsubsection{Loss Function}

To perform $\nu$-DBA, we define the geometric error loss function over optical flow and disparity 
\begin{equation}
\mathcal{L}_{f}=\sum_{\mathbf{r}\in R}\|\hat{\mathbf{F}}(\mathbf{r})-{\mathbf{F}}(\mathbf{r})\|_2,~
\mathcal{L}_{d}=\sum_{\mathbf{r}\in R}\|\hat{D}_{disp}(\mathbf{r})-{D}_{disp}(\mathbf{r})\|_2
\label{geo loss}
\end{equation}

Following common practice, we also add an Eikonal term on the sampled points to regularize SDF values in 3D space and sparsity loss to penalize the uncontrollable free surfaces:

\begin{equation}
 \mathcal{L}_{eik}=\sum_{\mathbf{x}\in \mathcal{X}}(\|\nabla \mathcal{S}_\theta(\mathbf{x})\|_2-1)^2 
\label{eik}
\end{equation}
\begin{equation}
 \mathcal{L}_{spa}=\sum_{\mathbf{x}\in \mathcal{X}}\exp(-\tau\cdot\|\mathcal{S}_\theta(\mathbf{x})\|) 
\label{spars}
\end{equation}
where $\tau$ is a hyperparameter to rescale the SDF value. 

To eliminate ambiguous occupancy of the close-range SDF field, an entropy regularization term is added
\begin{equation}
 \mathcal{L}_{ent}=f_{ent}(\hat{O}_c(\mathbf{r})), ~ f_{ent}(x)=-(x\ln x+(1-x)\ln(1-x))
\label{ent}
\end{equation}
where $\hat{O}(\mathbf{r})$ is the opacity of close-range SDF field along ray $\mathbf{r}$ that can be calculated as $\hat{O}(\mathbf{r}) = \sum_{i=1}^N T_i\alpha_i$.

The overall loss used for $\nu$-DBA is 
\begin{equation}
 \mathcal{L}_{\nu DBA}=\mathcal{L}_{f}+\lambda_1\mathcal{L}_{d}+\lambda_2\mathcal{L}_{eik}+\lambda_3\mathcal{L}_{spa}+\lambda_4\mathcal{L}_{ent}
\label{loss_gdba}
\end{equation}

\subsection{Road Surface Initialization} \label{rsi}
As discussed in \cite{guo2023streetsurf}, 
the disentanglement of close-range and distant view is an unsupervised and ill-posed problem with almost no constraints. 
The road surface initialization, wherein the close-range SDF field is pre-trained to possess a zero-level set approximately aligned with the road surface, is crucial for addressing this issue in driving scenes. To get the road surface, we leverage the corresponding information provided by the estimated optical flow. The corresponding points are triangulated with the preprocessed camera poses to obtain a 3D point cloud. Subsequently, we identify the plane with the maximal support within the point cloud by employing the Random Sample Consensus (RANSAC) algorithm. This best-fitting plane serves as the initial reference for the close-range SDF model.

\subsection{Voxel-based Sampling}
The sampling strategy is important for neural implicit field training. Recent works \cite{muller2022instant,mao2023ngel, long2022sparseneus, guo2023streetsurf} have proved that the octree structure can significantly improve sampling efficiency. As a result, we employ the octree to sample points along the rays in the close-range model. The octree is first initialized by the triangulated point cloud in subsection \ref{rsi} and is updated iteratively throughout the training of the close-range model. We uniformly sample $N$ points each voxel intersected by the ray. 
To further improve the sampling efficiency, we discard those sampling points whose SDF value is higher than the threshold after $60\%$ of the total number of training iterations. These sampling strategies ensure that the selected points are in proximity to the surface, thereby enhancing the model's accuracy.

\subsection{Color Rendering}
In this work, we consider color rendering optional as our geometric error provides sufficient guidance for the DBA task.
As discussed in \cite{zhu2022nice,zhu2023nicer}, when applying color-related loss and geometric-related loss together, the loss terms may conflict, degenerating the performance. 
Therefore, we decouple the color model from the geometry model and only learn the color model if novel view synthesis is required in addition. 






In the color model, we employ a shader for close-range view and a distant model that represents the scene hundreds or even thousands of meters away from the camera as well as the sky.

\begin{figure*}
\centering
\captionsetup[subfloat]{labelfont=scriptsize,textfont=scriptsize}
    \subfloat[Ours w/o ss]{
    \centering
        \begin{minipage}{0.32\linewidth}
            \centering
            \includegraphics[width=\linewidth]{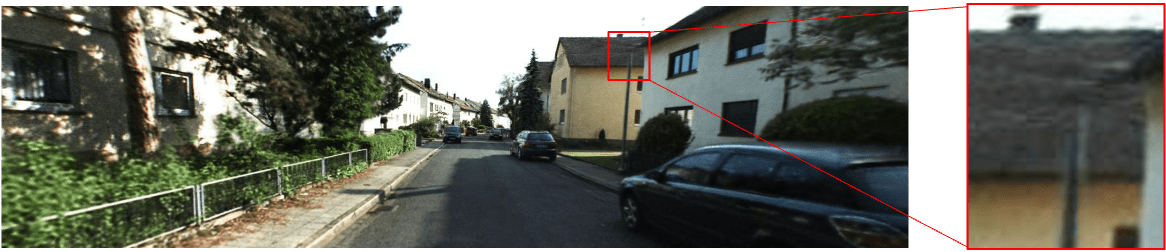}
            \end{minipage}
        }
    \subfloat[Ours]{
    \centering
        \begin{minipage}{0.32\linewidth}
            \centering
            \includegraphics[width=\linewidth]{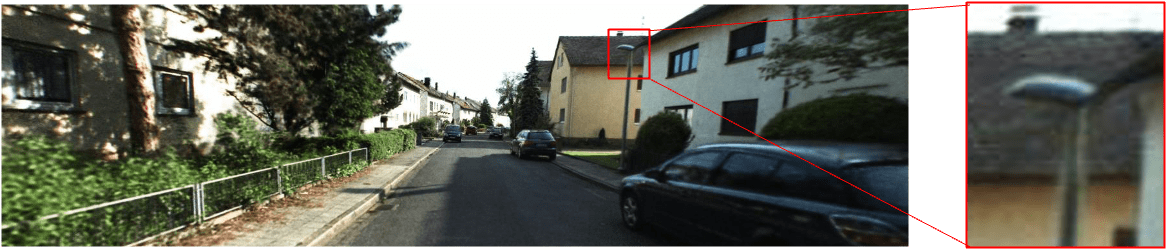}
            \end{minipage}
        }
        \subfloat[GT]{
        \centering
        \begin{minipage}{0.32\linewidth}
            \centering
            \includegraphics[width=\linewidth]{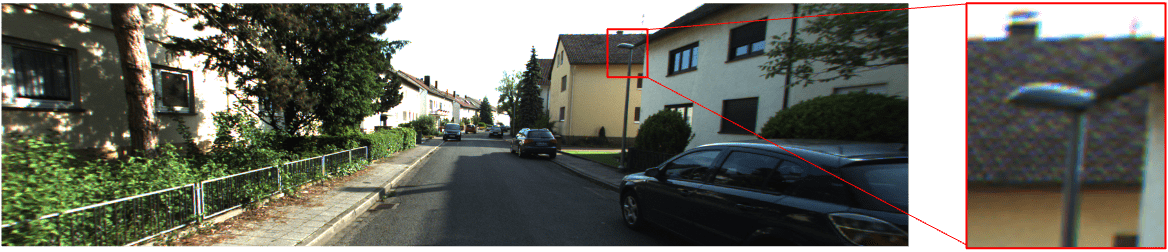}
            \end{minipage}
        }
\centering
\caption{\textbf{Ablation study on Self-supervised Fine-tuning.} Self-supervised (ss) fine-tuning improves the details of the model and achieves a better reconstruction.
}
\label{kitti_ablate}
\end{figure*}

\subsubsection{Close-Range Color Field}
Given point $\mathbf{x}$ and a view direction $\mathbf{v}$, the function of RGB color $\mathcal{C}$ can be defined as
\begin{equation}
\hat{\mathbf{c}} = \mathcal{C}_\theta(\mathbf{v}, \hat{\mathbf{z}})
\label{close model rgb}
\end{equation}
The feature vector $\hat{\mathbf{z}}$ is the output of the geometry decoder of the SDF model.

\subsubsection{Distant Model}

To handle unbounded scene rendering, we leverage a scene parameterization similar to \cite{guo2023streetsurf}. The distant model is a hyper 4D hash table. Given the 4D input $\mathbf{x}'$, the network directly outputs the density $\hat{\sigma}$ and RGB color $\hat{\mathbf{c}}$ of the sample point.
\begin{equation}
[\hat{\sigma}, \hat{\mathbf{c}}] = h^{d}_\theta(\mathbf{x}')
\label{distant model}
\end{equation}

To form a 4D input, we sample on cuboid shells and apply inverse cuboid warping. Samples of the distant-view model lie on cuboid shells. The cuboid shells are scaled proportionally from the cuboid shell of the close-range space with inverse-proportionally increasing scales. The warped point $\mathbf{x}'$ of the sample $\mathbf{x}$ on the $i$th cuboid shell can be expressed as 
\begin{equation}
\mathbf{x}'=\left[r_i\cdot\mathbf{x}, r_i\right],~ r_i=\frac{1}{(1-i/n)+(i/n)(1/r_{max})}
\label{inverse}
\end{equation}
where $n$ is the number of cuboid shells and $r_{max}$ is the scale of the largest cuboid shell relative to the close-range shell.

\subsubsection{Optimization}
Given a sample $x_i$, the color model outputs the RGB color value $\mathbf{c}$. Similar to the section. \ref{close vr}, the color $\hat{\mathbf{C}}$ for the given ray $\mathbf{r}$ is computed via numerical integration
\begin{equation}
\hat{\mathbf{C}}(\mathbf{r}) = \sum_{i=1}^N T_i\alpha_i \mathbf{c}_i
\label{render color}
\end{equation}


Then, we can optimize the color model with a simple photometric loss
\begin{equation}
\mathcal{L}_{photo}=\sum_{\mathbf{r}\in R}\|\hat{\mathbf{C}}(\mathbf{r})-{\mathbf{C}}(\mathbf{r})\|_2
\label{color loss}
\end{equation}



\section{Experiments}
In this section, we validate our method on various real-world outdoor datasets. We perform ablation study wrt. different geometry cues and provide qualitative and quantitative comparisons against state-of-the-art baselines and perform ablation study wrt. different geometry cues.

\subsection{Experiment Setup}

\subsubsection{Datasets}
In this work, we mainly focus on outdoor unbounded scenes with monocular input or stereo pair input. Thus
we consider 3 real-world datasets for evaluation:  KITTI-360 \cite{liao2022kitti}, Waymo \cite{sun2020scalability}, and a self-collected dataset from an autonomous driving mine truck with more unstructured views. 

\subsubsection{Baselines}
We compare our method to 1) state-of-the-art neural implicit surfaces methods: NeuS-facto \cite{nerfstudio} and StreetSurf \cite{guo2023streetsurf}. 2) neural radiance field methods: NeRFacto \cite{nerfstudio} and F2-NeRF \cite{wang2023f2nerf}.  3) state-of-the-art SLAM system: ORB-SLAM3 \cite{campos2021orb} and DROID-SLAM \cite{teed2021droid}.

\subsubsection{Metrics}
To evaluate the reconstruction accuracy, we report Accuracy [cm], Completion [cm], and Completion Ratio with a threshold of 20cm of the extracted meshes compared with ground truth LiDAR data. For a fair comparison, we remove unseen regions that are not inside any camera's viewing frustum and crop the meshes with the oriented bounding box of LiDAR data. For appearance, we evaluate the PSNR, SSIM, and LPIPS for rendered color images averaging on the training set and test set the same as \cite{guo2023streetsurf}. To evaluate the tracking accuracy, we use ATE [m].

\subsubsection{Implementation Details}

We implement our method in PyTorch \cite{NEURIPS2019_9015}. We use the cuboid hash-grids in nr3d \cite{guo2023streetsurf} as the hash table in scene representation. We sample 8192 rays per iteration and train our model for 20k iterations for about 30 minutes on a single NVIDIA RTX 4090 GPU.

\subsection{Ablation Study}
We verify whether the self-supervised fine-tuning of the optical flow model can improve the performance of the method and ablate the impact of different geometric cues on reconstruction quality. To avoid interference caused by incorrect pose, the comparison is conducted using ground truth trajectories. The data is averaged over four different scenes in the KITTI-360 dataset \cite{liao2022kitti}. 

\begin{table}[]
\begin{tabular}{llccc}
\toprule
 &  \makebox[0.15\linewidth]  & \makebox[0.15\linewidth]{Acc. $\downarrow$} & \makebox[0.15\linewidth]{Comp. $\downarrow$} & \makebox[0.15\linewidth]{Comp. Ratio $\uparrow$}\\ 
\midrule
\multirow{2}{*}{KITTI-360} & w/o ss & 40.92 & 25.38 & 73.42 \\
  & w/ ss  & \textbf{38.40} & \textbf{23.82} & \textbf{74.78}         \\ \midrule
\multirow{2}{*}{Waymo}    & w/o ss &  48.06 &  81.95  &    63.87 \\
  & w/ ss  &  \textbf{45.98}  &   \textbf{72.89}    &   \textbf{73.96}  \\ \bottomrule
\end{tabular}
\caption{\textbf{Ablation Study on Self-Supervised (ss) Fine-tuning.}}
    \label{ablatess}
\end{table}

\subsubsection{Self-Supervised Fine-tuning}
We compare the reconstruction quality before and after self-supervised fine-tuning and report metrics averaged over the KITTI-360 and Waymo datasets in TABLE \ref{ablatess}. Thanks to the narrower generalization gap, we observe that self-supervised fine-tuning improves the reconstruction accuracy. It is worth noting that the pre-trained models without (w/o) fine-tuning can also achieve relatively good results, but as shown in Fig. \ref{kitti_ablate}, self-supervised fine-tuning can improve the details of the model and achieve better reconstruction.

\begin{table}[]
\begin{tabular}{lccc}
\toprule
  \makebox[0.3\linewidth][l]{Geometry cues}  & \makebox[0.15\linewidth]{Acc. $\downarrow$} & \makebox[0.15\linewidth]{Comp. $\downarrow$} & \makebox[0.15\linewidth]{Comp. Ratio $\uparrow$}\\ 
\midrule
m-depth & 82.71 & 32.80 & 61.73 \\
m-depth+m-normal & 63.45 & 27.92 & 64.80 \\
flow & 36.79 & 25.65 & 71.14 \\
ss-flow  & 31.21 & 23.93 & 74.44 \\ 
stereo  & 27.17 & 20.27 & 78.71 \\
stereo+ss-flow  & \textbf{23.95} & \textbf{19.96} & \textbf{80.34}  \\ \bottomrule
\end{tabular}
\caption{\textbf{Ablation Study on Different Geometry Cues.} \texttt{ss-flow} denotes self-supervised optical flow, \texttt{m-} denotes monocular geometry cues.}
\label{ablategeo}
\end{table}

\subsubsection{Different Geometry Cues}
We now investigate the effectiveness of different geometric cues. To avoid interference from RGB cues, we do not use photometric loss in this experiment. We conducted monocular and stereo-pair experiments on the KITTI-360 dataset. Note that we also ablate monocular depth and normal cues with our model, and the loss function is the same as \cite{Yu2022MonoSDF}. As shown in TABLE \ref{ablategeo}, the results indicate that for monocular input, self-supervised flow outperforms monocular cues as expected. For stereo pair input, optical flow further helps stereo improve the reconstruction performance. Comparing monocular and stereo pair inputs, stereo pair provides better geometry, which is obvious with more information.

\begin{table}[]
\begin{tabular}{llccc}
\toprule
 &  & \makebox[0.15\linewidth]{Acc. $\downarrow$} & \makebox[0.13\linewidth]{Comp. $\downarrow$} & \makebox[0.13\linewidth]{Comp. Ratio $\uparrow$}\\ 
\midrule
\multirow{2}{*}{ss-flow} & w/ $\mathcal{L}_p$  & 38.40 & \textbf{23.82} & \textbf{74.78}\\
& w/o $\mathcal{L}_p$  & \textbf{31.21} & 23.93 & 74.44 \\ 
\midrule
\multirow{2}{*}{stereo+ss-flow} & w/ $\mathcal{L}_p$  & 26.42 & \textbf{19.67} & \textbf{80.53}\\
& w/o $\mathcal{L}_p$  & \textbf{23.95} & 19.96 & 80.34   \\ \bottomrule
\end{tabular}
\caption{\textbf{Ablation Study on Photometric Loss.}}
\label{ablatephoto}
\end{table}

\subsubsection{Photometric Loss}
In this experiment, we investigated the impact of photometric loss on reconstruction. As shown in TABLE \ref{ablatephoto}, photometric loss increases completeness at the cost of a significant decrease in accuracy. This result derives the conflict and ambiguity mentioned in \cite{zhu2022nice, zhu2023nicer}. We consider that the photometric loss hurts the geometry in the way of more parameters to learn, and illumination disturbance to color loss as found in sparse bundle adjustment literature \cite{campos2021orb}.

\subsection{Dense Bundle Adjustment} \label{dbae}

\subsubsection{Trajectory and Mapping} 
\begin{figure}
\centering
\captionsetup[subfloat]{labelfont=scriptsize,textfont=scriptsize}
    \subfloat[GT]{
        \centering
        \begin{minipage}{0.45\linewidth}
        \includegraphics[width=\linewidth,trim=4 4 4 4,clip]{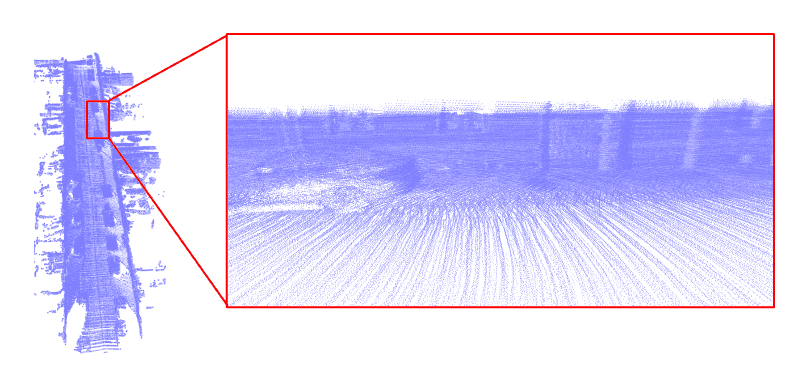}
    \end{minipage}
    }
    \subfloat[StreetSurf \cite{guo2023streetsurf}]{
    \centering
        \begin{minipage}{0.45\linewidth}
            \centering
            \includegraphics[width=\linewidth,trim=4 4 4 4,clip]{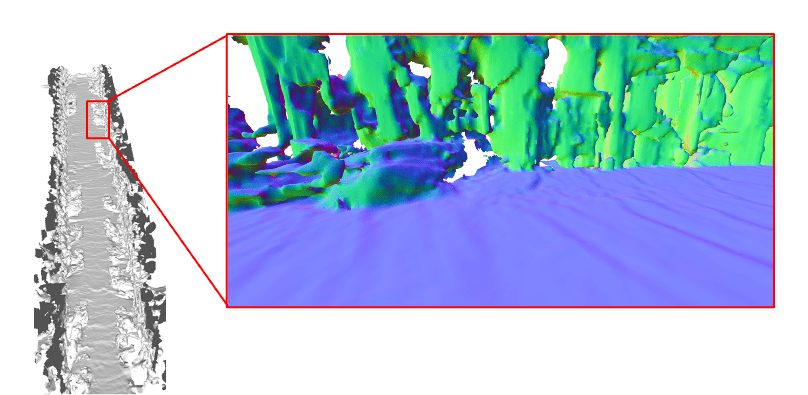}
            \end{minipage}
        }
        
    \subfloat[\textbf{Ours}]{
            \centering
            \begin{minipage}{0.45\linewidth}
            \includegraphics[width=\linewidth,trim=4 4 4 4,clip]{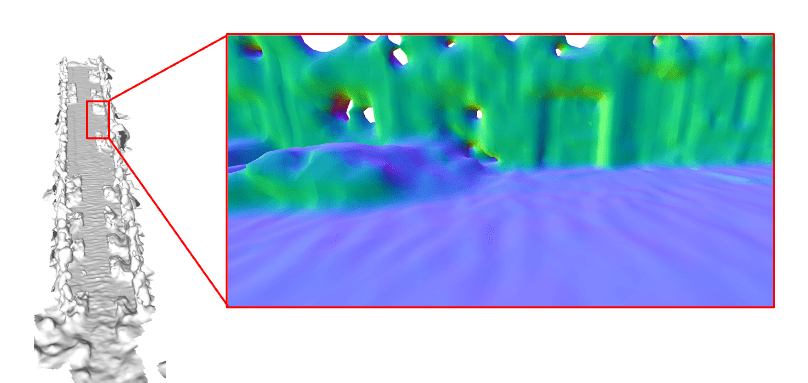}
        \end{minipage}
        }
    \subfloat[\textbf{Ours (w/ stereo)}]{
        \centering
        \begin{minipage}{0.45\linewidth}
        \includegraphics[width=\linewidth,trim=4 4 4 4,clip]{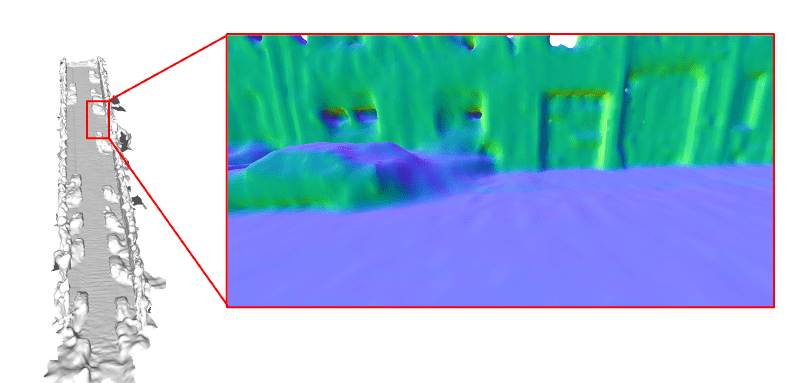}
    \end{minipage}
    }

\centering
\caption{\textbf{Reconstruction results on the KITTI-360 dataset.} Since no ground truth mesh is given, the ground truth LiDAR point cloud is displayed. 
}
\label{kitti_mesh}
\end{figure}

In this experiment, we demonstrate the effectiveness of our method as a dense geometric bundle adjustment with noisy camera poses. The results are averaged over four different scenes in the KITTI-360 dataset. We compare our method with the back-end of DROID-SLAM and StreetSurf. The input camera poses for all methods are obtained from the front-end of DROID-SLAM. ORB-SLAM3 is also utilized as a reference for tracking comparison. As shown in Table \ref{gdba_table}, we compare the accuracy of the trajectory as well as that of the resultant surface. All other methods outperform ORB-SLAM3, illustrating that dense geometric bundle adjustment is more effective than traditional sparse geometric bundle adjustment. DROID-SLAM is not as effective as the other two methods based on neural implicit surfaces, indicating the effectiveness of the 3D consistency provided by neural implicit representation. Compared with StreetSurf, our method employs a purely geometric BA without a photometric loss. Our method achieves the best performance in both tasks by eliminating color task disturbance and utilizing a 3D-consistent map parametrization. For qualitative comparison, as shown in Fig. \ref{kitti_mesh}, our method reconstructs more accurate and detailed surfaces.

\begin{table}[]
\begin{tabular}{lcccc}
\toprule
\makebox[0.13\linewidth][l]{Method} & \makebox[0.11\linewidth]{ATE $\downarrow$} & \makebox[0.11\linewidth]{Acc. $\downarrow$} & \makebox[0.11\linewidth]{Comp. $\downarrow$} & \makebox[0.15\linewidth]{Comp. Ratio $\uparrow$} \\ 
\midrule
ORB-SLAM3 \cite{campos2021orb} & 0.186 & - & - & - \\
DROID-SLAM \cite{teed2021droid} & 0.084  & 87.01 & 81.31 & 40.07 \\
StreetSurf \cite{guo2023streetsurf} & 0.078 & 75.52 & 38.63 & 56.51 \\
\textbf{Ours} &  0.073 & 41.63 & 30.62 & 62.48   \\ 
\textbf{Ours (w/ stereo)} & \textbf{0.071} & \textbf{29.77} & \textbf{25.35} &  \textbf{73.40}  \\ 
\bottomrule
\end{tabular}
\caption{\textbf{Quantitative comparison of bundle adjustment results.}}
\label{gdba_table}
\end{table}

\begin{figure}
\centering
\captionsetup[subfloat]{labelfont=scriptsize,textfont=scriptsize}
    \subfloat[F2-NeRF \cite{wang2023f2nerf}]{
    \centering
        \begin{minipage}{0.45\linewidth}
            \centering
            \includegraphics[width=\linewidth]{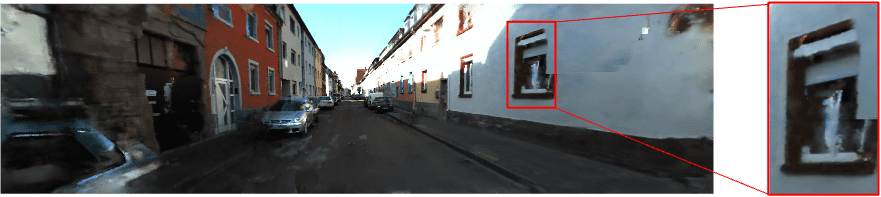}
            \end{minipage}
        }
    \subfloat[StreetSurf \cite{guo2023streetsurf}]{
    \centering
        \begin{minipage}{0.45\linewidth}
            \centering
            \includegraphics[width=\linewidth]{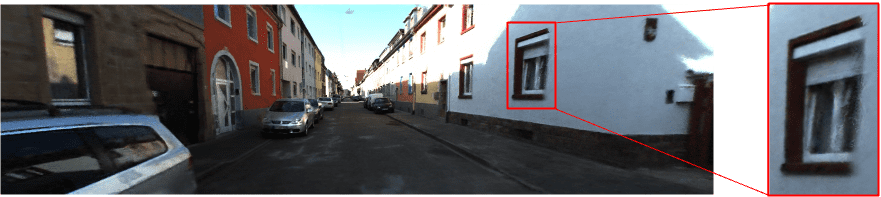}
            \end{minipage}
        }
        
    \subfloat[\textbf{Ours}]{
            \centering
            \begin{minipage}{0.45\linewidth}
            \includegraphics[width=\linewidth]{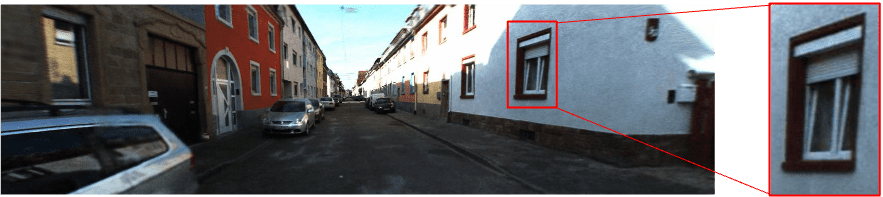}
        \end{minipage}
        }
    \subfloat[GT]{
        \centering
        \begin{minipage}{0.45\linewidth}
        \includegraphics[width=\linewidth]{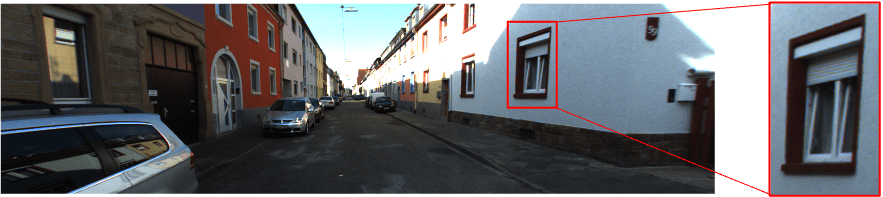}
    \end{minipage}
    }
        
\centering
\caption{\textbf{Rendering results on the KITTI-360 dataset.} 
}
\label{kitti_color}
\end{figure}

\begin{table*}[t]
\begin{tabular}{lccc|ccc}
\toprule
 \makebox[0.1\textwidth]  & \multicolumn{3}{c|}{\texttt{Appearance}} & \multicolumn{3}{c} {\texttt{Geometry}}\\
\midrule
 Method  & \makebox[0.11\textwidth]{PSNR $\uparrow$} & \makebox[0.11\textwidth] {SSIM $\uparrow$} & \makebox[0.11\textwidth] {LPIPS $\downarrow$}  & \makebox[0.11\textwidth] {Acc. $\downarrow$} & \makebox[0.11\textwidth] {Comp. $\downarrow$} & \makebox[0.11\textwidth] {Comp. Ratio $\uparrow$} \\ 
\midrule
NeRFacto \cite{nerfstudio} & 17.02 & 0.593 & 0.412 & - & - & - \\
F2-NeRF \cite{wang2023f2nerf} & 26.70 & 0.725 & 0.359 & - & - & - \\ 
NeuS-Facto \cite{nerfstudio} & 14.27  & 0.533 & 0.437 & 446.93 & 201.86 & 9.92 \\ 
StreetSurf \cite{guo2023streetsurf} & 30.25 & 0.887 & 0.342 & 62.56 & 74.84 & 70.19 \\ 
\textbf{Ours (ss-flow)} & \textbf{30.54} & \textbf{0.891} & \textbf{0.298} & \textbf{45.98} & \textbf{72.89} & \textbf{72.96} \\ 
\bottomrule
\end{tabular}
\caption{\textbf{Quantitative comparison of appearance and geometry on Waymo dataset.}}
\label{reprewaymo}
\end{table*}

\subsubsection{Trajectory and Colored Mapping}
In the previous subsection, we evaluate the results of $\nu$-DBA: camera tracking accuracy and reconstruction quality. In addition to $\nu$-DBA, our method, as a neural implicit scene representation, can render high-quality color images.

Continuing with the above experiment, we compare the two neural implicit-based methods, StreetSurf and our method, with other neural implicit scene representation methods: NeRFacto \cite{nerfstudio}, F2-NeRF \cite{wang2023f2nerf}, and NeuS-Facto \cite{nerfstudio}. For a fair comparison, the camera pose inputs for these three methods are also obtained from the front-end of DROID-SLAM, which is the same as Section \ref{dbae}. As shown in TABLE \ref{renderkitti}, both our methods with only self-supervised optical flow prior and with additional stereo pair outperform other methods. The qualitative experimental results are shown in Fig. \ref{kitti_color}, where our method achieves a sharper reconstruction than other methods.

\begin{table}[t]
\begin{tabular}{lccc}
\toprule
\makebox[0.2\linewidth][l]{Method}  & \makebox[0.18\linewidth]{PSNR $\uparrow$} & \makebox[0.18\linewidth] {SSIM $\uparrow$} & \makebox[0.18\linewidth] {LPIPS $\downarrow$}\\ 
\midrule
NeRFacto \cite{nerfstudio} & 16.89 & 0.580 & 0.421 \\
F2-NeRF \cite{wang2023f2nerf} & 22.24 & 0.781 & 0.365 \\ 
NeuS-Facto \cite{nerfstudio} & 16.23 & 0.550 & 0.432 \\ 
StreetSurf \cite{guo2023streetsurf}  & 24.37 & 0.816 & 0.329\\ 
Ours  & 25.14 & 0.830 & 0.324  \\ 
\textbf{Ours w/ stereo} & \textbf{25.93} & \textbf{0.840} & \textbf{0.316}  \\ 
\bottomrule
\end{tabular}
\caption{\textbf{Quantitative comparison of appearance on KITTI-360 dataset.}}
\label{renderkitti}
\end{table}

\subsection{Colored Mapping-only Bundle Adjustment} \label{color mapping}

The above experiment utilizes the noisy camera pose input, which means the accuracy of camera poses after BA will influence the experimental results. To eliminate interference caused by incorrect pose, we conduct a reconstruction experiment using ground truth camera pose input on the Waymo dataset. As shown in TABLE \ref{reprewaymo}, our method still performs better than other methods when the pose is accurate.

\begin{table}[]
\begin{tabular}{lccc}
\toprule
\makebox[0.2\linewidth][l]{Method} & \makebox[0.18\linewidth]{PSNR $\uparrow$} & \makebox[0.18\linewidth] {SSIM $\uparrow$} & \makebox[0.18\linewidth] {LPIPS $\downarrow$} \\
\midrule
StreetSurf \cite{guo2023streetsurf}& 26.92 & 0.766 & 0.389 \\
\textbf{Ours} & \textbf{27.33} & \textbf{0.823} & \textbf{0.385}\\
\bottomrule
\end{tabular}
\caption{\textbf{Quantitative comparison of appearance on Mine dataset.}}
\label{repremine}
\end{table}

    


    

\begin{figure}[]
\centering   
\captionsetup[subfloat]{labelfont=scriptsize,textfont=scriptsize}
    \subfloat{
    \rotatebox[origin=c]{90}{\scriptsize{GT}}
    \centering
        \begin{minipage}{0.9\linewidth}
            \centering
            \includegraphics[width=\textwidth]{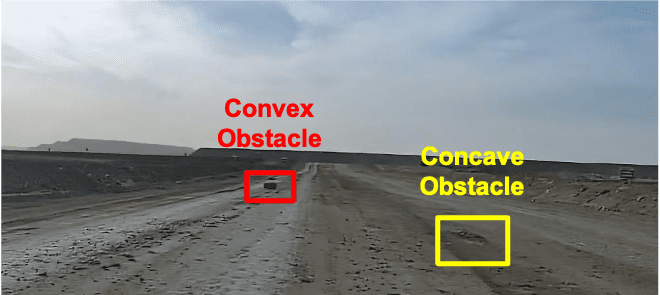}
        \end{minipage}
    }\vspace{1mm}

    \subfloat{
    \rotatebox[origin=c]{90}{\scriptsize{StreetSurf \cite{guo2023streetsurf}}}
        \begin{minipage}{0.9\linewidth}
            \centering
            \includegraphics[width=\textwidth,trim=4 4 4 4,clip]{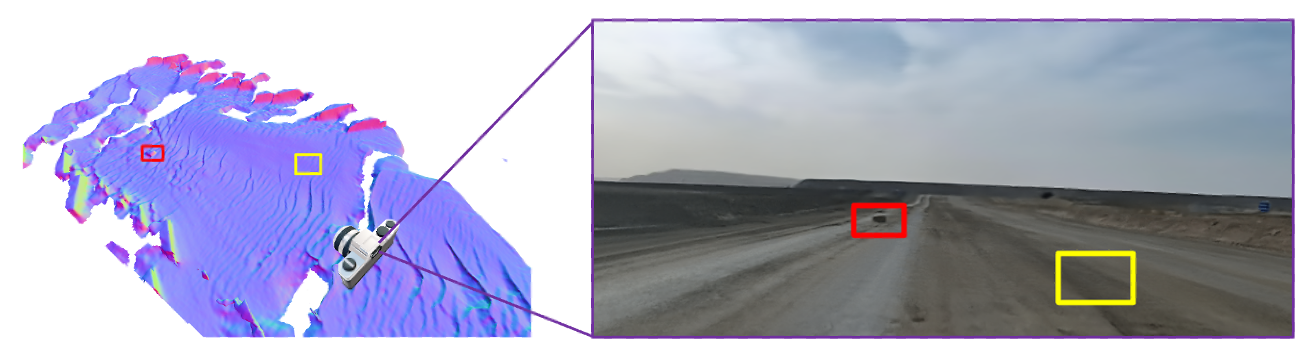}
        \end{minipage}
    }\vspace{1mm}

    \subfloat{
    \rotatebox[origin=c]{90}{\scriptsize{\textbf{Ours}}}
        \begin{minipage}{0.9\linewidth}
            \centering
            \includegraphics[width=\textwidth,trim=4 4 4 4,clip]{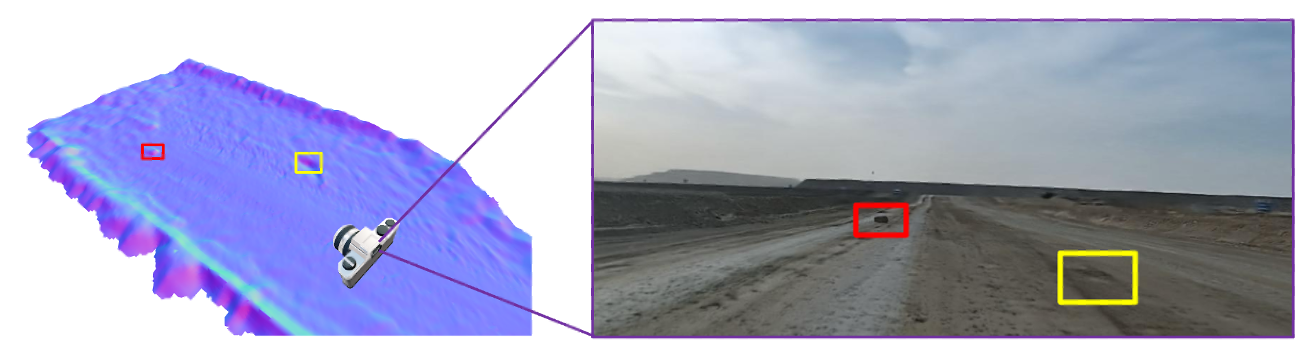}
        \end{minipage}
    }

    \caption{\textbf{Rendering and reconstruction results on the Mine dataset.} The convex obstacle is about 30cm high and the concave obstacle is about 20cm deep.}
    \label{meshmine}
\end{figure}

KITTI-360 and Waymo are both driving scene datasets in street view. To validate the generalization of our method, we utilize a monocular outdoor dataset on a more unstructured view self-collected from an autonomous driving mine truck. Since the dataset lacks ground truth geometry data and camera poses, we only quantitatively analyzed the rendering results in TABLE \ref{repremine}. The rendering results and reconstruction results are shown in Fig. \ref{meshmine}. On this challenging dataset, our method rendered higher-fidelity RGB images and extracted more precise meshes with challenging obstacles.  

All the experiment results above demonstrate that our method achieves better performance on camera tracking and reconstruction thanks to our $\nu$-DBA and eliminating color task disturbance.

\section{Conclusion}

In this paper, we propose $\nu$-DBA, a novel geometric dense BA framework that utilizes a 3D neural implicit surface representation as the map parametrization. This framework simultaneously optimizes the neural implicit map surface and the camera trajectory poses by minimizing geometric error derived from dense optical flow across consecutive frames, thereby bridging the 3D neural implicit representation with geometric error minimization to enhance the accuracy of dense bundle adjustment.

In addition, we investigate the effects of photometric error and other neural geometric prior on the accuracy of surface reconstruction and novel view synthesis. Moreover, we refine the flow model through per-scene self-supervision for better performance. Compared with other bundle adjustment methods and neural implicit reconstruction methods, our method achieves better performance in pose optimization and reconstruction.

\bibliographystyle{IEEEtran}
\bibliography{IEEEabrv,refs}

\end{document}